\documentclass{article}

\usepackage{PRIMEarxiv}

\usepackage[utf8]{inputenc} 
\usepackage[T1]{fontenc}    
\usepackage{hyperref}       
\usepackage{url}            
\usepackage{booktabs}       
\usepackage{amsfonts}       
\usepackage{nicefrac}       
\usepackage{microtype}      
\usepackage{xcolor}         
\usepackage{amsmath}
\usepackage{amssymb}
\usepackage{mathtools}
\usepackage{amsthm}
\usepackage{bbm}
\usepackage{subfig}
\usepackage{natbib}
\definecolor{ForestGreen}{RGB}{34,139,34}
\DeclareMathOperator*{\argmin}{\arg\!\min}

\DeclareMathOperator*{\argmax}{\arg\!\max}

\theoremstyle{plain}
\newtheorem{theorem}{Theorem}[section]
\newtheorem{proposition}[theorem]{Proposition}

\theoremstyle{definition}

\newtheorem{assumption}[theorem]{Assumption}
\theoremstyle{remark}

\newcommand{\loss}{ESR}
\DeclareMathOperator{\sgn}{sgn}

\newcommand\numberthis{\addtocounter{equation}{1}\tag{\theequation}}

\title{Asymptotically Optimal Regret for Black-Box Predict-then-Optimize
}

\author{%
  Samuel Tan \\
  Cornell University\\
  Ithaca, NY 14850 \\
  \texttt{sst76@cornell.edu} \\
  \And
  Peter I. Frazier \\
  Cornell University\\
  Ithaca, NY 14850 \\
  \texttt{pf98@cornell.edu} \\
}

\begin{document}
\maketitle

\begin{abstract}
We consider the predict-then-optimize paradigm for decision-making in which a practitioner (1) trains a supervised learning model on historical data of decisions, contexts, and rewards, and then (2) uses the resulting model to make future binary decisions for new contexts by finding the decision that maximizes the model's predicted reward. This approach is common in industry. Past analysis assumes that rewards are observed for all actions for all historical contexts, which is possible only in problems with special structure. Motivated by problems from ads targeting and recommender systems, we study new black-box predict-then-optimize problems that lack this special structure and where we only observe the reward from the action taken. We present a novel loss function, which we call Empirical Soft Regret (\loss{}), designed to significantly improve reward when used in training compared to classical accuracy-based metrics like mean-squared error. This loss function targets the regret achieved when taking a suboptimal decision; because the regret is generally not differentiable, we propose a differentiable “soft” regret term that allows the use of neural networks and other flexible machine learning models dependent on gradient-based training. In the particular case of paired data, we show theoretically that optimizing our loss function yields asymptotically optimal regret within the class of supervised learning models. We also show our approach significantly outperforms state-of-the-art algorithms on real-world decision-making problems in news recommendation and personalized healthcare compared to benchmark methods from contextual bandits and conditional average treatment effect estimation.
\end{abstract}


\section{Introduction}
Predict-then-optimize is an approach to contextual decision-making problems that separates prediction of context- and decision-dependent costs from optimization.
Such separation is common in practice and widely used in industry.
A rapidly growing line of work in predict-then-optimize focuses on loss functions to use in model training \citep{wilder2019melding, elmachtoub2022smart, mandi2022decision}. The key insight in this work is that traditional accuracy-based metrics may not yield the best downstream decision after the optimization step; alternative loss functions that target optimizing the downstream decision often perform better. 

This past analysis and associated methodology, however, relies on a special form for the objective function: It assumes that the historical data is of the form $(w,c)$, where $w$ is a context variable and $c$ determines the cost of all decisions $x$ through a known function with input $x$ and $c$. 
This fails to capture many realistic settings in which the form of the objective is not known and must also be estimated, such as ads targeting and recommender systems. Fine-tuning in large language models is a more modern example in which the model, when presented with a prompt, must choose a completion with the highest reward, with the reward model being unknown.

For example, consider maximizing advertising clicks by choosing which advertisement to show each impression, where each impression is associated with contextual features such as the user's browsing history and demographic information. Supposing that data was collected from a randomized policy to eliminate hidden confounding, one can fit a model to predict the click-through rate as a function of the shown advertisement and context; when presented a context, the model chooses the advertisement that maximizes the predicted click-through rate.
Critically, when we train this model, we only know whether the user clicked on the ad shown. We do not know whether they would have clicked if we had shown another ad. This prevents using existing methods from predict-then-optimize.

To fill in this gap in the predict-then-optimize literature, we consider the following \textit{black-box} predict-then-optimize problem arising when 
the reward is only observed for the action taken and the form of the objective function is unknown.
This proceeds in two stages. The first stage is prediction: one trains a supervised learning model on historical data $(y_i, x_i, w_i)_{i=1}^n,$ where $y_i$ is a continuous outcome or reward observed from taking an action $x_i$ based on context $w_i$, and we assume that for any $x,w$, there exists a ground truth model $y = f(x,w)$. Specifically, empirical risk minimization (ERM), with some choice of loss, is used to optimize over the space of model parameters, yielding a model $\hat{f}_\theta(x,w)$. 
The second stage is optimization: action $\hat{x}(w) \in \argmax_{x \in \mathcal{X}} \hat{f}_\theta(x,w)$ is taken when the algorithm is presented with context $w$, with the hope that the true outcome $y$ associated with taking action $\hat{x}(w)$ with context $w$ is near-optimal, that is, $f(\hat{x}(w), w) \approx \max_x f(x,w)$. 
This is similar to the predict-then-optimize setting, except that the form of the objective is unknown and we are unable to reconstruct $f(x,w_i)$ for $x\ne x_i$.
We focus on binary actions, though our approach is likely to generalize to multiple discrete actions.

We show using theory and experiments that ERM using the classical mean-squared error (MSE) loss function can fail to achieve low regret, even when the amount of available data is large and there is a model in our model class that would provide lower regret.
To illustrate this difficulty, 
suppose training with MSE found a model $\hat{f}_{MSE}$ with the smallest regret. 
Suppose we then modify $f(x,w)$ by shifting the ``level'', i.e., by defining the new model $f'(x,w) = f(x,w) + a_w$ which adds a value $a_w$ that depends on $w$ alone, and regenerate the training data. This would not change the optimal decision or the regret incurred by the previously fitted model. It would, however, change its MSE. If the model class is not flexible enough to include the new model $f'$, then ERM with MSE will typically select a different model that models the shift, providing lower MSE, but has higher regret.
In other words, the ML model may be ``distracted'' by focusing on modeling the contextual features, when it should be focused on modeling the effect of the decision $x$. While we focus on the MSE, MAE or other accuracy-based measures would face the same issue.


We provide a novel solution:
we propose an alternative loss function that targets minimizing the regret incurred when using $\hat{f}_\theta$ for decision-making. This loss function, which we call the {\it Empirical Soft Regret} (ESR), is parameterized by a positive constant $k$ and  is defined as 
\[ \frac{1}{n} \sum_{i=1}^n \frac{\|f(x_i,w_i) - f(x_{n(i)},w_{n(i)}) \|}{1 + \exp\left(k \sgn\{f(x_i,w_i) - f(x_{n(i)},w_{n(i)})\} \left(\hat{f}_\theta(x_i,w_i) - \hat{f}_\theta(x_{n(i)},w_{n(i)}) \right)\right)}, \]
where datapoints $i,n(i)$ form a pair such that 
$x_i \ne x_{n(i)}$ and $w_i,w_{n(i)}$ are as close as possible. The form of this loss function is such that as $k$ grows large and assuming $w_{n(i)} \approx w_i$, this approximates 

\[ \frac{1}{n} \sum_{i=1}^n  \left ( \max_{x \in \{0,1\} } f(x, w_i) - f(\argmax_{x \in \{0,1\}} \hat{f}_\theta (x, w_i), w_i) \right), \]

which is the exact form of the empirical regret in the binary case. We use the "soft" version to make the loss function differentiable, as it is not for $k = \infty$, i.e. the exact regret.
Therefore this method preserves the benefits of separating prediction and optimization (for any black box ML model that supports custom differentiable loss functions), but additionally yields lower regret.

Bringing theory to bear, we show that under general function classes, minimizing the \loss{} loss function yields asymptotically optimal regret when counterfactuals are observed; that is, it achieves regret that goes to 0 with the number of datapoints.

We complement this theory via numerical experiments showing that training on \loss{} outperforms  training on MSE as well as other standard approaches in causal machine learning on a causal machine learning benchmark dataset in personalized healthcare and on a news recommender dataset from Yahoo.

The rest of the paper is structured as follows. \S\ref{sec:related} overviews related literature. \S\ref{sec:model} presents our mathematical setting and \S\ref{sec:method} presents our proposed loss function. \S\ref{sec:theory} presents our main theoretical result. \S\ref{sec:exp} demonstrates the effectiveness of our method on experiments in both conditional average treatment effect estimation and contextual bandits. Finally, \S\ref{sec:conc} concludes and outlines future work.

\section{Related Work}
\label{sec:related}

\begin{table}
\centering
\begin{tabular}{ |c|c| } 
 \hline
Area  & Key Differences \\ 
 \hline
 Predict-then-optimize  & Full vs. bandit feedback \\ 
  Contextual bandits  & Sequential/adaptive vs. batch/one-shot \\ 
 Causal ML/CATE estimation  & Estimator accuracy vs. regret minimization \\ 
 Learning to rank  &  Permutation vs. single item as output \\
 \hline
\end{tabular}
\caption{Summary of related work.}
\label{tab:rel}
\end{table}

Table \ref{tab:rel} summarizes the most closely related literature areas. Below, we discuss them in more detail.
\subsection{Predict-then-optimize}
The literature considering problems most similar to this work is the budding field of predict-then-optimize, also called decision-focused learning. Like this work, the literature there constructs alternative loss functions when training models for downstream decision-making. Works in this space assume the following general framework. Assume the objective function to be maximized has a known form $f(x,c)$, where $x$ is the decision variable which must lie in some set $\mathcal{X}$ and $c$ is an uncertain parameter. Training data is given of the form $\{(w_i, c_i)\}_{i=1}^n$, where some parametric model $\hat{c}(w)$ is fit by regressing $c \sim w$. Then, when presented with a context $w$, the approach outputs decision $\argmax_{x \in \mathcal{X}} f(x, \hat{c}(w))$ and the regret is the loss in objective due to using a prediction for the true context-dependent value for $c$, $c(w)$, i.e. 
\[\max_{x\in \mathcal{X}} f(x, c(w)) - f( \argmax_{x \in \mathcal{X}} f(x, \hat{c}(w)), c(w)).\]

This setting crucially differs from ours because, in the above model, the form of $f(x,c)$ is known. This implies that for each historical datapoint $(w_i, c_i)$, the outcome corresponding to each alternate decision $x$ is known, i.e. $f(x, c_i)$ is a known quantity for each $x\in \mathcal{X}$; all counterfactuals are known. This makes, for instance, maximizing the objective within the training data trivial; simply take $\argmax_{x \in \mathcal{X}} f(x, c_i)$ for each point $i$. This is in contrast to our setting, in which we only observe the output of the response function for a single action and context. Because we do not observe all possible actions for each context, we must build a regression model $y \sim x, w$ that can predict counterfactual outcomes, and maximizing the outcome in the training data is not immediate.

Several notable papers consider the setting outlined in the previous paragraph.
\cite{elmachtoub2022smart} considers exclusively linear objectives where $f(x,c)  = c^\top x$ and $\mathcal{X}$ is a nonempty, compact, and convex set. In this setting, they derive a convex surrogate loss for the decision loss using duality. \cite{wilder2019melding} considers combinatorial optimization problems where $\mathcal{X}$ is a discrete set enumerating the feasible solutions. For cases when $f$ corresponds to linear optimization and submodular maximization problems, the authors construct a related problem for which it is possible to differentiate through the $\argmax$ operator, allowing gradient-based optimization.

To our knowledge, the work most related to ours in this space is \cite{mandi2022decision}, which considers applying learning-to-rank methods to combinatorial optimization problems. They consider general $f$, whose form is known to the decision-maker, and for which $\mathcal{X}$ is a discrete set of integer feasible solutions. The paper considers alternatives to MSE in regressing $c \sim w$, including pairwise and listwise losses. However, their paper (1) does not consider a loss which explicitly aims to minimize regret and (2)  only demonstrates empirical results for these alternative loss functions without any theoretical justification, while we provide theory justifying our \loss{} loss.

\subsection{Contextual Bandits}

Contextual bandits are sequential decision-making  problems where an agent takes actions $x_t$ when presented with contexts $w_t$ and earns rewards $r_t$, with the aim of maximizing cumulative reward up to a time horizon $T$. Solving contextual bandits is supported by maintaining a model of predicted rewards, and updating the model as data is observed. This highlights a crucial difference between contextual bandits and our setting: in our setting the model used to generate decisions is fixed after training. 

Within contextual bandits, the most closely related work is the direct method, also known as value-based learning. In this approach, the contextual bandit algorithm first learns a Q function (a model of the arm's reward under a given context), and then executes the greedy policy with respect to the Q function \citep{dudik2014doubly}. The loss function usually used to learn the Q function is MSE. In contrast, we propose an alternative and better-performing loss function. (There also exists theoretical work which establishes that value-based learning in contextual bandits is provably competitive with the optimal policy \citep{brandfonbrener2021offline}.)


Thus, one way to use our loss function is to improve the performance of value-based learning in contextual bandits. Our experiments on the news recommender dataset demonstrates this. (This dataset is often used for testing contextual bandit algorithms \citep{li2010contextual, semenov2022diversity}.) Indeed, our experiments compare standard value-based learning using MSE against value-based learning using our loss function. These experiments show that our loss function improves performance. 

\subsection{Causal ML/CATE Estimation}

In causal machine learning, data of the form $(y_i, x_i, w_i)$ is often observed, where $x_i$ is the treatment indicator, $w_i$ are contextual features, and $y_i$ is the outcome. The potential outcomes framework assumes each individual has two potential outcomes $y_i(1), y_i(0)$, and we only observe $y_i(x_i)$. The goal is to estimate the conditional average treatment effect (CATE) defined as $\mathbb{E}[y(1) - y(0) | x]$, toward which the literature 
constructs 
loss functions 
for training 
black-box machine learning models. Metalearners comprise many of the most common methods in this space; these are machine learning workflows which include custom loss functions, but also include constructing additional training signals, data splitting, and multiple ERM steps. Moreover, theoretical results in this work guarantee the quality of the CATE estimate and its relationship with the quality of the ML predictions \citep{kunzel2019metalearners, kennedy2023towards, nie2021quasi}. In contrast, we focus on choosing the highest-reward treatment, whereas these metalearners focus on estimating CATE with high-accuracy (e.g., low MSE); this aligns with policy learning, the problem of learning a mapping from an individual's context to a treatment.

One may interpret our work as “decision-focused” CATE estimation. One can frame CATE/policy learning within the black-box predict-then-optimize paradigm. In particular, the analogous method to the direct method from contextual bandits would be to (1) first use the S-learner \citep{kunzel2019metalearners}, which forms a model for the outcomes as a function of the treatment and covariates and (2) choose the (binary) treatment T which maximizes the predicted outcome. Our contribution is then a better loss function to use with the S-learner (as opposed to MSE) that yields better treatment decisions. Section \ref{sec:ihdp} shows that we outperform other metalearners on a standard benchmark task.

\subsection{Learning to Rank}
In learning to rank, we train a model to ingest a query $w$ and a list of items $x$ (represented by features) and output a permutation of the items $\pi(x_1, x_2, \ldots)$. The goal is for the permutation of the items to reflect the order of the items’ relevance to the query. The training data generally consists of queries $q$, items $x$ for each query $q$, and relevance scores $y$ for each item $x$ and query $q$.

To produce such a model, approaches include the pointwise, pairwise, and listwise approaches; the pointwise approach \citep{liu2009learning} most closely resembles our black-box predict-then-optimize paradigm. The pointwise approach first regresses a relevance score $y$ on a query $w$ and item $x$, where the training data contains multiple items x per query w. The output is the permutation in decreasing order of predicted score.

In one special case, this setting matches our black-box predict-then-optimize paradigm. This special case is the case where a permutation’s quality is judged by the top-1 metric \citep{niu2012top}. (The top-1 metric gives a reward of 1 if the first item in the permutation is relevant ($y=1$) and 0 if not.) Using the top-1 metric with the pointwise approach would correspond to outputting any permutation with the first item being the item with the highest predicted score and considering this to be an optimal decision if this item is relevant. Note that, of course, lists generally contain more than two elements, and so this setting extends beyond the binary action case.

Outside of this special case, our setting differs from learning to rank. Taking $f(x,w)$ to be the relevance score of item $x$ for query $w$, our setting returns the argmax of $\hat{f}_\theta(x,w)$, i.e., a single item, while learning to rank returns a permutation. When the relevance metric depends on the full permutation, it cannot be computed from the single item that we return.

One can also treat $x$ as the full permutation, and $f(x,w)$ as a listwise relevance score of permutation $x$ in response to query $w$. But in this setting, one observes $f(x,w)$ for all of the permutations of the items in the training data, not just one, i.e. this is the standard predict-then-optimize setting, not the black-box setting we consider. Our approach could be used to study learning to rank problems with bandit feedback, where we only observe a score for the permutation actually shown (and we are not willing to make structural assumptions that would let us compute scores for permutations not shown).
The training of LLMs also often relies on using human preference data to fine-tune its capabilities to ensure safe and helpful output. \cite{rafailov2023direct} show that directly optimizing over pairs of human preference data yields better or equal performance to methods based on reinforcement learning with human feedback (RLHF) which rely on learning a reward model. 
\section{Model} 
\label{sec:model}

The mathematical model we assume is as follows. Suppose a stakeholder must make a binary decision $x \in \{0,1\}$. They are provided a set of contexts $\{w\}$ that describe the setting in which they make this decision; in an advertising example, for instance, this may be comprised of features that describe a user, e.g. device used, browsing history, recorded interests, etc. Given a decision $x$ and context $w$, the stakeholder receives a reward $f(x, w)$. The goal of the stakeholder is to choose $x$ to maximize expected reward, where the expectation is taken over contexts $w$. To do this, the user is presented with historical data of the form $(y_i, x_i, w_i)$ where $y_i = f(x_i, w_i)$ and the datapoints $i$ are not necessarily collected sequentially. They use the this set of data to train a single model $\hat{f}_\theta(x, w)$, where $\theta$ represents some parameterization of $\hat{f}_\theta$; this captures, for example, the linear weights in ordinary least squares, or the entire set of weights and biases in a deep neural network. After training, the model is fixed and is used to make decisions $\argmax_{x \in \mathcal{X}} \hat{f}_\theta(x, w)$ when presented with a set of context $\{w\}$. Such a setting is reasonable when there are engineering limitations to regularly updating a ML model. This is in contrast to a contextual bandit setting, an online learning problem in which the model is continually updated as observations of the outcome are made.

\section{Methodology}
\label{sec:method}

Our methodology revolves around constructing an alternative loss function that targets the regret incurred by taking suboptimal decisions. We define regret as 

\[ \mathrm{R}_\theta(w):= \max_{x \in \{0,1\}} f(x, w) - f(\argmax_{x \in \{0,1\}} \hat{f}_\theta (x, w), w) \]

The interpretation of regret is that we are comparing the outcome corresponding to the ground truth best possible action against the outcome corresponding to the action chosen by a given model. Importantly, both actions are evaluated under the true response function $f$, as we do not care about $\hat{f}_\theta$'s ability to approximate $f$, only it's ability to generate good decisions $\argmax_{x} \hat{f}_\theta (x, w).$ This is in contrast to the generic loss function used in regression problems, mean squared error (MSE).

In particular, consider the empirical MSE loss:
\begin{equation*}L_{MSE}(f, \hat{f}_\theta) := \frac{1}{n} \sum_{i=1}^n \left( f(x_i, w_i) - \hat{f}_\theta (x_i, w_i) \right)^2.\end{equation*}
Observe that in training with MSE, there is no difference in interpretation between the decision variables $x_i$ and contextual features $w_i$, despite the fact that $x_i$ represents a decision to be made. In other words, there is no consideration of the downstream impact of the mean squared error on regret.

Ideally, a machine learning model could directly minimize the empirical regret:
\[ \frac{1}{n} \sum_{i=1}^n R_{\theta}(w_i) = \frac{1}{n} \sum_{i=1}^n \left ( \max_{x \in \{0,1\} } f(x, w_i) - f(\argmax_{x \in \{0,1\}} \hat{f}_\theta (x, w_i), w_i) \right). \]
However, this is difficult to actually use as a loss function because the only dependence of the loss on the parameter $\theta$ is through an $\argmax$, making any gradient-based optimization nontrivial to apply.

To motivate our proposed loss function, we first rewrite the empirical regret as 
\[\frac{1}{n} \sum_{i=1}^n \mathbbm{1}\{\sgn \{f(1,w_i) - f(0,w_i)\} \neq \sgn\{\hat{f}_\theta(1,w_i) - \hat{f}_\theta(0,w_i)\}\} \|f(1,w_i) - f(0,w_i)\|,\]
as (1) we incur regret only when the predicted sign of the difference between $\hat{f}_\theta(1,w_i) - \hat{f}_\theta(0,w_i)$ is wrong and (2) the magnitude of the regret, if incurred, is the magnitude of the true difference between taking actions 1 and 0. In this form, the loss function remains highly nonsmooth with respect to the parameter $\theta$ due to the indicator function. Instead, consider the ``soft'' version which replaces the indicator function with a sigmoid function:

\[\frac{1}{n} \sum_{i=1}^n \frac{\|f(1,w_i) - f(0,w_i) \|}{1 + \exp\left(k \sgn\{f(1,w_i) - f(0,w_i)\} \left(\hat{f}_\theta(1,w_i) - \hat{f}_\theta(0,w_i) \right)\right)}.\]

\begin{figure}
\centering
\includegraphics[scale = 0.5]{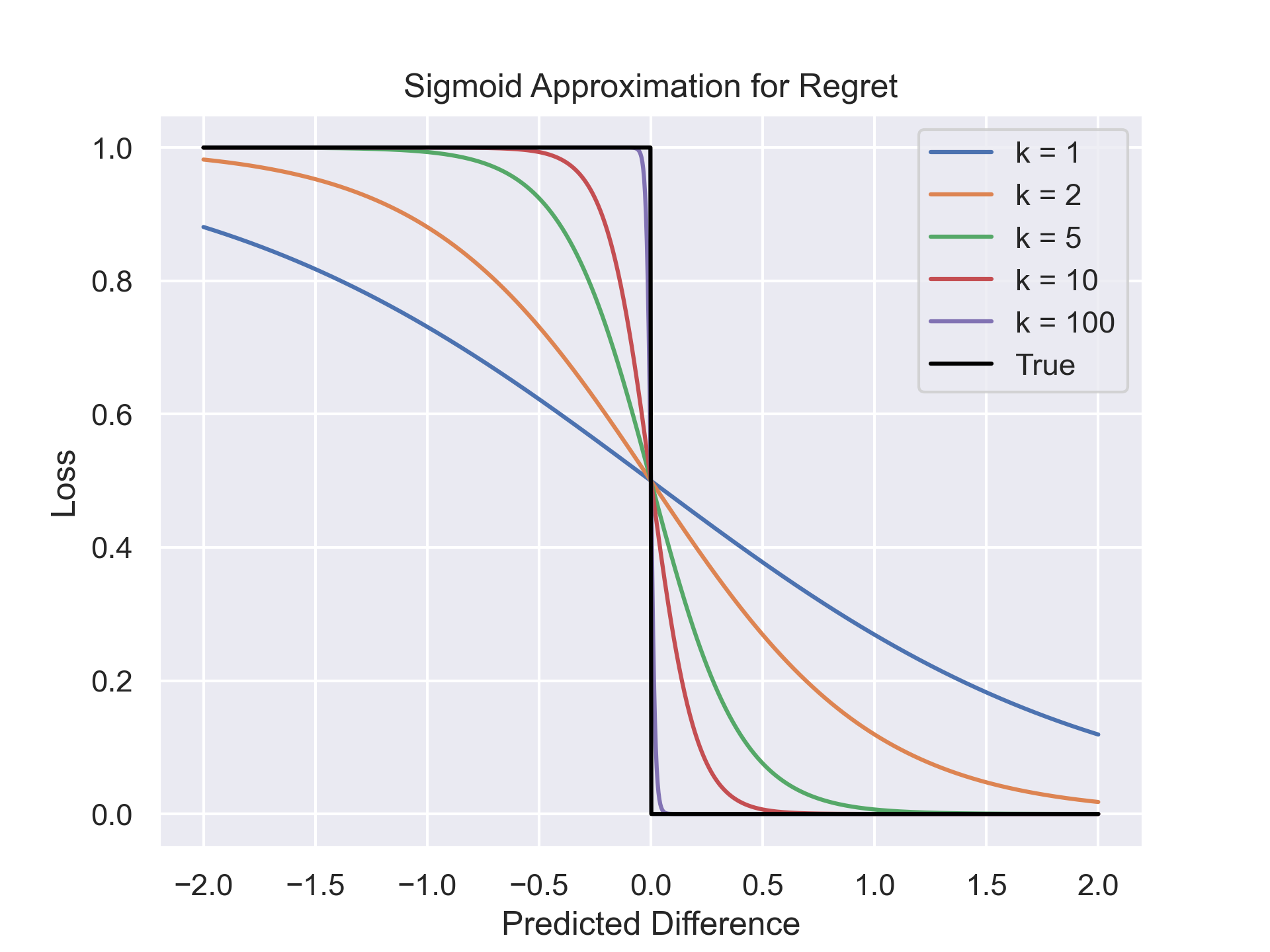}
\caption{Dependence of \loss{} function on $k$, for fixed $w$ and $f(1,w) > f(0,w)$.}
\label{fig:sig}
\end{figure}

Here we see the loss is smooth with respect to $\theta$, where the level of smoothness is controlled by $k$. Figure \ref{fig:sig} illustrates how $k$ affects the smoothness of the loss function for a fixed $w$. Therefore using this loss in lieu of the actual regret enables the use of autodifferentation and gradient-based optimization such as in neural networks.

This leads us to our \loss{} loss function, which accomodates the setting that the dataset does not contain both $x=0,1$ for each $w$. First, denote $n(i)$ as the index of the nearest neighbor for point $i$ such that the \textit{decision} $x$ is different, i.e. $x_{n(i)} \neq x_i$. Specifically, we define the nearest neighbor for datapoint $i$ as 
\[n(i) \in \argmin_{\substack{j=1, \ldots, n \\ {j\neq i} \\ {x_i \neq x_j}}} \| w_i - w_{j} \|_2. \]

Then our loss function is
\[  L_{\loss,k}(f, \hat{f}_\theta) := \frac{1}{n} \sum_{i=1}^n \frac{\|f(x_i,w_i) - f(x_{n(i)},w_{n(i)}) \|}{1 + \exp\left(k \sgn\{f(x_i,w_i) - f(x_{n(i)},w_{n(i)})\} \left(\hat{f}_\theta(x_i,w_i) - \hat{f}_\theta(x_{n(i)},w_{n(i)}) \right)\right)}, \]
where again $k$ is a parameter which controls the smoothness of the loss function. The use of the nearest neighbor approximates the behavior of a perfectly matched datapoint, i.e. observing both $x=0,1$ for $w_i$. In settings when $f(x,w)$ is well-behaved in $w$, this should be a reasonable assumption.

A key observation about using the \loss{} function is that the output of the model trained on the \loss{} loss function does not necessarily yield an interpretable output $\hat{f}_\theta(x,w)$ for $y_i$. Instead, it is only the case that the function $\hat{f}_\theta (x,w)$ has the correct maximizer, relative to $f(x,w)$.



\section{Theory}
\label{sec:theory}

We state the following theorem on the performance of using \loss{} loss function in regression settings. For the result, we assume that the data is perfectly paired so that we always observe both $x = 0, 1$ for the same context $w$. We require the following assumptions:

\begin{assumption} There exists some upper bound $\Delta$ to the difference in objective between the actions:
\[ \sup_{w}  |f(1,w) - f(0,w)| \leq \Delta .\]
\end{assumption}

We also require a condition on the function class $\hat{f}_\theta$ in terms of its covering number $N_\Theta(\alpha)$.

\begin{assumption}
    Suppose the function space $\hat{f}_\theta$ defined by the set of $\theta \in \Theta$ has covering number which satisfies $N_\Theta(\alpha) \in O(\frac{1}{\alpha^2})$.
\end{assumption}

This assumption covers common flexible machine learning models such as support vector machines \citep{guo1999covering} and some neural networks \citep{bartlett2017spectrally}.

Finally, we require the following assumption on the asymptotic distribution of $\hat{f}_\theta(1,w) - \hat{f}_\theta(0,w)$:

\begin{assumption}
$  \lim_{n \to \infty} \sup_\theta \mathbb{P} \left(|\hat{f}_\theta(1,w) - \hat{f}_\theta(0,w) | < n^{-1/4} \right) = 0,$\label{assump:3}
\end{assumption}
which ensures the sigmoid function is asymptotically never evaluated at $x=0$.

\begin{theorem}
Define $\theta^*$ as the minimizer of the expected regret within $\Theta$, and define $\hat{\theta}_{\loss,n}$ as the minimizer of the \loss{} function over $n$ datapoints. Then for $k \geq  n^{1/4} \log n$ and under the above assumptions, 

\[ | \mathbb{E}_{w \sim \mathcal{W}} R_{\theta^*}(w) - \mathbb{E}_{w \sim \mathcal{W}} R_{\hat{\theta}_{\loss,n}}(w)  | \overset{n \to \infty}{\rightarrow} 0 \]

with probability exponentially going to 1.
\end{theorem}

See the proof in the appendix.


\section{Experiments}
\label{sec:exp}

This section contains results from experiments in a semi-synthetic benchmark dataset based on the Infant Health and Development Program (IHDP) and a real news recommender dataset from Yahoo, where we compare the performance of state-of-the-art models with that of \loss{}-trained models.

\subsection{IHDP Dataset}
\label{sec:ihdp}

We evaluate our method on a standard benchmark for evaluating conditional average treatment effect (CATE) estimators. The dataset is based on the Infant Health and Development Program
(IHDP), which measured the impact of specialist home visits on future cognitive test scores for $n=747$ (608 control/139 treatment) children. Each individual child is represented by 25 covariates. We follow the procedure used in \cite{hill2011bayesian}, \cite{johansson2016learning},\cite{shalit2017estimating}, \cite{johansson2022generalization}, which uses the log-linear simulated outcome implemented as setting “A” in the
NPCI package \citep{dorie2016npci}. We use 1000 replications with a 70:30 train/test split.

For comparison, we pit a model trained on the \loss{} loss against several state-of-the-art CATE estimators in the literature: the S-learner, the T-learner \citep{kunzel2019metalearners}, the R-learner \citep{nie2021quasi}, and the DR-learner \citep{kennedy2023towards}. The CATE estimate is used to produce a decision by taking the sign of the estimated CATE, i.e. $\hat{x} = \sgn{\widehat{CATE}(w)}$. Each CATE estimator uses a two-layer neural network.

As this dataset is semi-synthetic, the counterfactual outcome is given and so regret can be calculated straightforwardly. Table \ref{tab:ihdp} shows the performance of the learners compared against the model trained with the \loss{} loss. We see that the \loss{} loss-trained model handily outperforms the other metalearners in regret. Table \ref{tab:ihdp} also shows the regret incurred when varying $k$ (which controls the smoothness of the loss). The performance varies but is generally stable in the $[5,100]$ range.

\begin{table}[t]
\label{tab:yahoo}
\vskip 0.15in
\begin{center}
\begin{tabular}{cccc}
\toprule
 Learner & Regret \\
\midrule
S   & [1.02, 1.36] \\
T & [0.70, 0.97] \\
R    & [2.81, 3.19] \\
DR    & [0.77, 1.21]      \\
\loss{}     & \textcolor{ForestGreen}{[0.35, 0.43]} \\
\bottomrule
\end{tabular}
\quad
\begin{tabular}{cc}
\toprule
$k$&Regret\\
\midrule
1 & [0.52, 0.64] \\
5 & [0.39, 0.49] \\
10 & [0.35, 0.44] \\
50 & [0.37, 0.45] \\
100 & [0.40, 0.49] \\
\bottomrule
\end{tabular}
\end{center}
\caption{(L) CTR (95\% CI) for test regret over 1000 replications. The ESR loss was computed using $k=25$ (R) Comparison of regret vs. $k$, 95\% CI over 1000 replications.
\label{tab:ihdp}
}
\end{table}

\subsection{News Recommender}
\label{sec:yahoo}

To evaluate the performance of our method on real-world data, we use a news recommender dataset from Yahoo! (R6A from Yahoo Webscope). This dataset is comprised of over 45 million user visits to the Yahoo Today Module over a ten day period in May 2009. On the Today Module, each user was presented with an article chosen uniformly at random from a pool of 20 articles, and Yahoo recorded whether the user clicked. Each user is associated with a set of 5 features (plus a constant feature), which were derived from embedding the actual user-level data to obscure personally identifiable information. The machine learning task is to predict whether a user will click, given the action space of the twenty articles $\{ a_1, \ldots, a_{20}\}$ and the user-level features. As we focus on the binary setting, we only select two articles at random from each pool, filtering out only the user visits which were presented one of those two articles. 

The data is given as ten separate files, one for each day the experiment was ran. In each day, there are between 30 and 60 different pools of articles shown to each user, where the pool of articles change over time. The number of users associated with each pool varies between tens and hundreds of thousands. In this dataset, there do exist cases where one user is shown multiple articles in the pool, corresponding to a situation where a user logs on to the Yahoo Today module multiple times within a short time frame. In other words, there are perfect matches in this dataset where we observe outcomes for which the contextual features are the same but the actions (and potentially response) are different.

Nevertheless, there are cases where a user is only presented with one article, necessitating calculation of nearest neighbors. In this case, the default method is to calculate all pairwise distances and choose the minimum. When several neighbors are equidistant, we sample one uniformly at random.

For these experiments, we use a standard fully-connected neural network with two hidden layers for all machine learning models. As the click data is binary, we use a sigmoid activation for the final layer, though we still use a regression loss function to be consistent with our theory and previous experiment. For comparison, we compare our method with the direct method (value-based learning) approach, which maintains a model for $f(x,w)$ that minimizes MSE and greedily selects the action which maximizes the estimate. As the decision space is binary, we can also naturally adapt the causal metalearners to this setting as well (in this case, the $S$-learner is equivalent to the direct method).

\subsubsection{Evaluation}
For each article pool, we select 70\% of the users as the training set and the remaining 30\% as the test set. Because we do not have click data for both articles for each user, we must use another metric to gauge performance. Given a policy $\pi$ that deterministically maps from $w$ to action $x$, we use the expected click-through rate of the policy $\pi$, where the expectation is over features $w$.
To evaluate this, we use \textit{off-policy} evaluation, a technique common in contextual bandits \citep{agarwal2019lec}, leveraging the fact that the dataset was collected under a known random policy (uniform random decisions).

The quantity to estimate is the value of policy $\pi$, denoted $V(\pi).$ This is defined as $ V(\pi) = \mathbb{E}_{w \in W} \left[ y|w, \pi(w)\right],$ where $W$ is the distribution of contexts from which the observed $w_i$ are drawn.

To estimate this quantity, we only look at the datapoints $(y_i, x_i, w_i)$ such that our model's outputted decision $\pi(w_i)$ matches the observed action $x_i$, and compute the average performance out of these points. Specifically, we define our estimator $\hat{V}(\pi)$ as 

\begin{equation}
\hat{V}(\pi) := \frac{\sum_{i=1}^n \mathbbm{1}\{\pi(w_i) = x_i\} y_i }{\sum_{i=1}^n \mathbbm{1}\{\pi(w_i) = x_i\}}
\label{eq:unbiased}
\end{equation}

We have the following result (proof in the Appendix) on the consistency of this estimator:

\begin{proposition}
The estimator $\hat{V}(\pi)$ is consistent for $V(\pi)$ as $n \to \infty$.
\end{proposition}

\begin{figure}[ht]
\vskip 0.2in
\begin{center}
\centerline{\includegraphics[scale = 0.5]{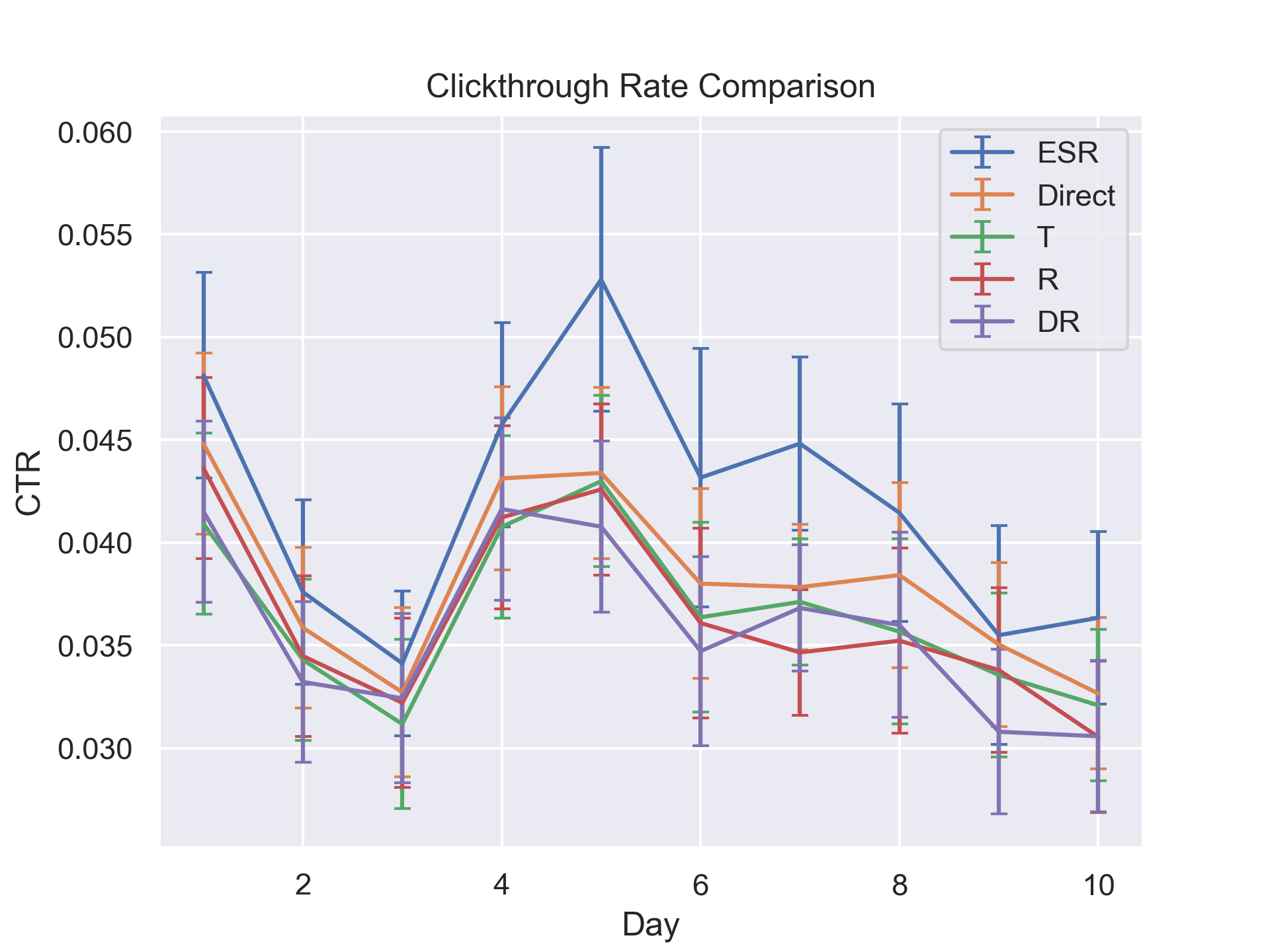}}
\caption{95\% CI for CTR of \loss{} loss-trained model vs. other methods over ten days.}
\label{fig:yahoo}
\end{center}
\vskip -0.2in
\end{figure}

Table \ref{tab:yahoo} and Figure \ref{fig:yahoo} summarize the results. Across all days the model trained on the \loss{} loss has higher mean CTR than the other methods, and its overall CTR is statistically significantly higher. This suggests that the model trained on the \loss{} loss is indeed doing a better job of capturing the dependence of the click-through rate on the article choice.

\begin{table}[t]
\label{tab:yahoo}
\vskip 0.15in
\begin{center}
\begin{small}
\begin{sc}
\begin{tabular}{cccccc}
\toprule
 Day & \loss{} & Direct & T & R & DR \\
\midrule
1   & [4.31, 5.32] & [3.99, 4.97] & [3.65, 4.53] & [3.94, 4.78] & [3.74, 4.56] \\
2 & [3.31, 4.21] & [3.10, 4.07] & [3.04, 3.82] & [3.00, 3.90] & [2.92, 3.72] \\
3    & [3.06, 3.76]& [2.90, 3.65] & [2.71, 3.53] & [2.80, 3.64] & [2.78, 3.71] \\
4    & [4.08, 5.07]& [3.83, 4.80] & [3.63, 4.52] & [3.61, 4.64] & [3.70, 4.62]      \\
5     & [4.64, 5.92] & [3.82, 4.86] & [3.88, 4.72] & [3.87, 4.65] & [3.67, 4.49] \\
6      & [3.69, 4.95] & [3.30, 4.31] & [3.41, 4.02] & [3.09, 3.84] & [3.32, 4.04] \\
$7$      & [4.06, 4.90] & [3.31, 4.26] & [3.41, 4.02] & [3.09, 3.84] & [3.09, 4.11]\\
$8$      & [3.62, 4.67] & [3.31, 4.38] & [3.12, 4.02] & [3.09, 3.96] & [3.09, 4.11]\\
$9$      & [3.02, 4.08] & [2.96, 4.04] & [2.96, 3.76] & [2.91, 3.85] & [2.62, 3.54]\\
$10$      & [3.21, 4.06] & [2.85, 3.68] & [2.84, 3.58] & [2.69, 3.41] & [2.66, 3.46]\\
\midrule
All      & \textcolor{ForestGreen}{[4.11, 4.45]} & [3.72, 4.04] & [3.57, 3.84] & [3.56, 3.84] & [3.50, 3.79]\\
\bottomrule
\end{tabular}
\end{sc}
\end{small}
\caption{CTR (95\% CI) for \loss{} Loss Trained NN vs. other offline contextual bandit methods. The mean CTR for each day is highest for the model trained on the \loss{} loss. The performance improvement is statistically significant over all ten days.}
\label{tab:yahoo}
\end{center}
\vskip -0.1in
\end{table}

\section{Conclusion and Future Work}
\label{sec:conc}

In this paper, we tackled the problem of training machine models which are used as a basis for decision-making. Instead of mean squared error, we proposed an alternative \loss{} loss function which targets the learning of the model towards minimizing the regret incurred by choosing the objective-maximizing decision. We justified our approach theoretically in that optimizing the \loss{} loss function yields asymptotically optimal regret within a model class. We also provided numerical results on a standard semi-synthetic benchmark and real-world data that our approach outperforms standard methods within contextual bandit and CATE estimation problems.

There are several future directions that stem from this work. An immediate next step is to develop methodology for more general actions spaces beyond binary $x$, including discrete $x$ and eventually continuous $x$. Likewise, we would like to extend our theory to capture unpaired data (which naturally would require some assumptions on the structure of $f$ with respect to $w$).

We are interested in exploring ways to apply similar methods to other machine learning models, especially reinforcement learning. Likewise, we also see potential to extend our methodology to learning from preferences in the fine-tuning of large language models as in \cite{rafailov2023direct}. 

Finally, we are improving the ability to make decisions based on data. When this is used by those whose goals are aligned with the broader society, then this should tend to improve societal outcomes, but our results could also be used by those whose goals are not so-aligned.


\bibliographystyle{plainnat}
\bibliography{bib}

\appendix

\section{Proofs}

\subsection{Proof of Theorem 1}

\begin{proof}

We first (re)define the following terms.
\[R_\theta(w) := \mathbbm{1}\{\sgn \{f(1,w) - f(0,w)\} \neq \sgn\{\hat{f}_\theta(1,w) - \hat{f}_\theta(0,w)\}\} \|f(1,w) - f(0,w)\|,\]
which is the regret incurred by approximating $f$ with $\hat{f}_\theta$ on a given context $w$.

Next, define the approximate ``soft'' regret as

\[R'_{\theta,k}(w) := \frac{\|f(1,w) - f(0,w) \|}{1 + \exp\left(k \sgn\{f(1,w) - f(0,w)\} \left(\hat{f}_\theta(1,w) - \hat{f}_\theta(0,w) \right)\right)} \]

Now we define additional terms:

\begin{itemize}
\item Expected regret:
\[ r(\theta):= \mathbb{E}_{w \sim \mathcal{W}} \left[R_\theta(w) \right]\]
\item Expected ``soft'' regret:
\[ s_k(\theta) := \mathbb{E}_{w \sim \mathcal{W}} [\hat{R}_{\theta, k}(w)],\]
where each $w_i$ are sampled from $\mathcal{W}$.
\item Empirical ``soft'' regret:
\[ \hat{s}_{n,k}(\theta) := \frac{1}{n} \sum_{i=1}^n R'_{\theta,k} (w_i).\]
\end{itemize}
Finally, define
\begin{itemize}
\item $\theta^* \in \argmin_\theta r(\theta)$ 
\item $\theta^{s} \in \argmin_\theta s_k(\theta)$ 
\item $\hat{\theta}^s_{n} \in \argmin_\theta \hat{s}_{n,k}(\theta)$ .
\end{itemize}

Now let us proceed with bounding $\|r(\theta^*) - r(\hat{\theta}^s_n) \|$.

We first apply the triangle inequality

\begin{align}
    \|r(\theta^*) - r(\hat{\theta}^s_n) \| \leq \|r(\theta^*) - s_k(\theta^s) \| + \|s_k(\theta^s) - \hat{s}_{n,k}(\hat{\theta}^s_n) \| + \|\hat{s}_{n,k}(\hat{\theta}^s_n) - s_k(\hat{\theta}^s_n) \|+ \| s_{k}(\hat{\theta}^s_n)  -r(\hat{\theta}^s_n)\| .
    \label{eq:tri1}
\end{align}

We first consider the third term, which is the difference between the empirical and expected soft regret. For notational simplicity, denote $\delta(w) := f(1,w) - f(0,w)$ and $\delta_\theta(w) = \hat{f}_\theta(1,w) - \hat{f}_\theta(0,w)$.

We have that 
\begin{align*} \|\hat{s}_{n,k}(\hat{\theta}^s_n) - s_k(\hat{\theta}^s_n) \| &\leq \sup_\theta  \|\hat{s}_{n,k}(\theta) - s_k(\theta) \| \\
&= \sup_\theta  \| \frac{1}{n} \sum_{i=1}^n \frac{\|\delta(w_i) \|}{1 + \exp\left(k \sgn\{\delta(w_i)\} \left(\delta_\theta(w_i) \right)\right)}   - \mathbb{E}  \left[ \frac{\|\delta(w) \|}{1 + \exp\left(k \sgn\{\delta(w)\} \left(\delta_\theta(w) \right)\right)} \right] \|
\end{align*}

Let $N_\Theta(\alpha)$ denote the covering number for the (potentially infinite) set of functions $\{\hat{f}_\theta\}_{\theta \in \Theta}$, which is defined as follows. Define a covering $C_{\Theta}(\alpha)$ as a subset of $\{\hat{f}_\theta\}$ such that for any $\hat{f}_\theta$, there exists $c \in C_{\Theta}(\alpha)$ such that $\|\hat{f}_\theta - c\|_\infty \leq \alpha$. Then define the covering number $N_\theta(\alpha)$ as the cardinality of the smallest such covering; call that covering (in terms of $\theta)$ $\tilde{\Theta}(\alpha).$

Then, using the covering $\tilde{F}_\Theta(\alpha),$ we proceed with the following inequalities. Define $\tilde{\theta} \in \Theta(\alpha)$ as the corresponding element which is closest to $\hat{f}_\theta$. We have that $    \sup_{\theta \in \Theta} \|s_k(\theta) - \hat{s}_{n,k}(\theta) \|$ is equal to

\begin{align*}
& \sup_\theta  \| \frac{1}{n} \sum_{i=1}^n \frac{\|\delta(w_i) \|}{1 + \exp\left(k \sgn\{\delta(w_i)\} \left(\delta_\theta(w_i) \right)\right)}   - \mathbb{E}  \left[ \frac{\|\delta(w) \|}{1 + \exp\left(k \sgn\{\delta(w)\} \left(\delta_\theta(w) \right)\right)} \right] \| \\
    =& \sup_{\theta \in \Theta} \|\mathbb{E}_w \frac{\|\delta(w) \|}{1 + \exp\left(k \sgn\{\delta(w)\} \left(\delta_{\tilde{\theta}}(w) \right)\right)} - \mathbb{E}_w \frac{\|\delta(w) \|}{1 + \exp\left(k \sgn\{\delta(w)\} \left(\delta_\theta(w) \right)\right)}\\
    &+ \frac{1}{n} \sum_{i=1}^n \frac{\|\delta(w_i) \|}{1 + \exp\left(k \sgn\{\delta(w_i)\} \left(\delta_\theta(w_i) \right)\right)} -
    \frac{1}{n} \sum_{i=1}^n \frac{\|\delta(w_i) \|}{1 + \exp\left(k \sgn\{\delta(w_i)\} \left(\delta_{\tilde{\theta}}(w_i) \right)\right)} \\
    &+ \frac{1}{n} \sum_{i=1}^n \frac{\|\delta(w_i) \|}{1 + \exp\left(k \sgn\{\delta(w_i)\} \left(\delta_{\tilde{\theta}}(w_i) \right)\right)}   - \mathbb{E}  \left[ \frac{\|\delta(w) \|}{1 + \exp\left(k \sgn\{\delta(w)\} \left(\delta_{\tilde{\theta}}(w) \right)\right)} \right] \|\\
    \leq& \sup_{\theta \in \Theta} \|\mathbb{E}_w \frac{\|\delta(w) \|}{1 + \exp\left(k \sgn\{\delta(w)\} \left(\delta_{\tilde{\theta}}(w) \right)\right)} - \mathbb{E}_w \frac{\|\delta(w) \|}{1 + \exp\left(k \sgn\{\delta(w)\} \left(\delta_\theta(w) \right)\right)} \| \\
    &+  \sup_{\theta \in \Theta} \| \frac{1}{n} \sum_{i=1}^n \frac{\|\delta(w_i) \|}{1 + \exp\left(k \sgn\{\delta(w_i)\} \left(\delta_\theta(w_i) \right)\right)} -
    \frac{1}{n} \sum_{i=1}^n \frac{\|\delta(w_i) \|}{1 + \exp\left(k \sgn\{\delta(w_i)\} \left(\delta_{\tilde{\theta}}(w_i) \right)\right)} \| \\
    &+ \sup_{\theta \in \Theta} \| \frac{1}{n} \sum_{i=1}^n \frac{\|\delta(w_i) \|}{1 + \exp\left(k \sgn\{\delta(w_i)\} \left(\delta_{\tilde{\theta}}(w_i) \right)\right)}   - \mathbb{E}  \left[ \frac{\|\delta(w) \|}{1 + \exp\left(k \sgn\{\delta(w)\} \left(\delta_{\tilde{\theta}}(w) \right)\right)} \right] \|\\
    \leq & \sup_{\theta \in \Theta} \|\mathbb{E}_w \frac{\|\delta(w) \|}{1 + \exp\left(k \sgn\{\delta(w)\} \left(\delta_{\tilde{\theta}}(w) \right)\right)} - \mathbb{E}_w \frac{\|\delta(w) \|}{1 + \exp\left(k \sgn\{\delta(w)\} \left(\delta_\theta(w) \right)\right)} \| \\
    &+  \sup_{\theta \in \Theta} \| \frac{1}{n} \sum_{i=1}^n \frac{\|\delta(w_i) \|}{1 + \exp\left(k \sgn\{\delta(w_i)\} \left(\delta_\theta(w_i) \right)\right)} -
    \frac{1}{n} \sum_{i=1}^n \frac{\|\delta(w_i) \|}{1 + \exp\left(k \sgn\{\delta(w_i)\} \left(\delta_{\tilde{\theta}}(w_i) \right)\right)} \| \\
    &+ \numberthis \label{step3} \max_{\theta' \in \tilde{\Theta}(\alpha)} \| \frac{1}{n} \sum_{i=1}^n \frac{\|\delta(w_i) \|}{1 + \exp\left(k \sgn\{\delta(w_i)\} \left(\delta_{\theta'}(w_i) \right)\right)}   - \mathbb{E}  \left[ \frac{\|\delta(w) \|}{1 + \exp\left(k \sgn\{\delta(w)\} \left(\delta_{\theta'}(w) \right)\right)} \right] \|\\
\end{align*}

Now we observe that the function $g(x) := \frac{1}{1 + \exp(kx)}$ has Lipschitz constant $k/4$. Furthermore, by definition of $\alpha$-covering, $\| f_{\tilde{\theta}} - \hat{f}_\theta\|_\infty \leq \alpha$ which implies $| \left( f_{\tilde{\theta}}(1,w) - f_{\tilde{\theta}}(0,w) \right) - \left( f_{\theta}(1,w) - f_{\theta}(0,w) \right) | \leq 2 \alpha$ by the triangle inequality. 

Therefore for any $w$,
\begin{align*} \|  \frac{1}{1 + \exp\left(k \sgn\{\delta(w)\} \left(\delta_{\tilde{\theta}}(w) \right)\right)}  - \frac{1}{1 + \exp\left(k \sgn\{\delta(w)\} \left(\delta_{\theta}(w) \right)\right)}  \| &\leq \frac{k}{4} \|  \delta_{\tilde{\theta}}(w)  - \delta_{\theta}(w) \| \\
&\leq \frac{2k\alpha}{4}\\
&= \frac{k\alpha}{2}.
\end{align*}

For the last term of \eqref{step3}, we first note that 
each $R'_{\theta,k}(w)$ is a random variable bounded between $0$ and $\Delta$, and so we may apply Hoeffding + union bound.

So for each $\theta' \in \tilde{\Theta}(\delta),$ by Hoeffding's we have

\[ \mathbb{P}\left(\| \mathbb{E}_w R'_{\theta',k}(w) - \frac{1}{n} \sum_{i=1}^n  R'_{\theta',k}(w_i)
 > t \right) \leq 2\exp\left(-2nt^2/\Delta^2 \right).\]

 Then the probability that the max of the $\theta'$'s deviations is larger than $t$ is the event that the union of the events that each $\theta'$'s deviation is larger than $t$. Therefore by the union bound, 

 \[ \mathbb{P}\left(\max_{\theta' \in \tilde{\Theta}(\delta)}\| \mathbb{E}_w R'_{\theta',k}(w) - \frac{1}{n} \sum_{i=1}^n R'_{\theta',k}(w_i) 
 > t \right) \leq 2 N_\Theta(\alpha) \exp\left(-2nt^2/\Delta^2 \right).\]

 Combining the above, we have that $\sup_{\theta \in \Theta} \|s_k(\theta) - \hat{s}_{n,k}(\theta) \| \leq k\alpha + t$ with probability at least $1-2 N_\Theta(\alpha) \exp\left(-2nt^2/\Delta^2 \right)$. 

Now consider the fourth term of \eqref{eq:tri1}, which has to do with how well $s_k$ approximates $r$. Expanding out the notation, we have 

\[ \| s_{k}(\hat{\theta}^s_n)  -r(\hat{\theta}^s_n)\| \leq \sup_\theta \Big| \mathbb{E}_{w \in \mathcal{W}} \frac{\|\delta(w) \|}{1 + \exp\left(k \sgn\{\delta(w)\} \delta_{\theta}(w) \right)} - \mathbbm{1}\{\sgn \{\delta(w)\} \neq \sgn\{\delta_{\theta}(w) \}\} \|\delta(w)\|  \Big|,\]
which we can first bound by 
\[ \Delta \Big| \mathbb{E}_{w \in \mathcal{W}} \frac{1}{1 + \exp\left(k \sgn\{\delta(w)\} \delta_{\theta}(w) \right)} - \mathbbm{1}\{\sgn \{\delta(w)\} \neq \sgn\{\delta_{\theta}(w) \}\}   \Big|,\]

which we can further separate into the positive and negative $\delta(w_i):$

Note that the case that $\delta(w_i) = 0$ doesn't matter because the regret will always be 0 on those $w_i$. Therefore the regret incurred will always be less than if all $|\delta(w_i)| > 0$.

\begin{align*} \Delta & \Big|  \int_{w: \delta(w) > 0} \frac{1}{1 + \exp\left(k  \delta_{\theta}(w_i) \right)} - \mathbbm{1}\{\sgn\{\delta_{\theta}(w_i) \} \neq 1\} p(w)dw   \\ &+ \int_{w: \delta(w) < 0} \frac{1}{1 + \exp\left(-k  \delta_{\theta}(w_i) \right)} - \mathbbm{1}\{\sgn\{\delta_{\theta}(w_i) \} \neq -1\}p(w) dw \Big|,\
\end{align*}
which is bounded above by
\begin{align*}
    \Delta & \Big|  \int_{w: \delta(w) > 0} \frac{1}{1 + \exp\left(k  \delta_{\theta}(w) \right)} - \mathbbm{1}\{\sgn\{\delta_{\theta}(w) \}\neq 1\} p(w) dw\Big| \\
    &+\Delta  \Big| \int_{w: \delta(w) < 0} \frac{1}{1 + \exp\left(-k  \delta_{\theta}(w_i) \right)} - \mathbbm{1}\{\sgn\{\delta_{\theta}(w_i) \}\neq -1\} p(w) dw\Big|.
\end{align*}

However, note that for $\delta_{\theta}(w) \neq 0$, $1 - \frac{1}{1 + \exp\left(k  \delta_{\theta}(w) \right)} = \frac{1}{1 + \exp\left(-k  \delta_{\theta}(w) \right)}$ and $1 - \mathbbm{1}\{\sgn\{\delta_{\theta}(w) \}\neq1\} = \mathbbm{1}\{\sgn\{\delta_{\theta}(w) \}\neq-1\}$. When $\delta_{\theta}(w) = 0,$ then both terms in the summation are equal (to $\frac{1}{2}$).

Therefore the terms within the absolute values are negations of each other, and we can recombine to yield 

\[ \Delta  \mathbb{E}_{w \in \mathcal{W}} \left[ \Big|  \frac{1}{1 + \exp\left(k  \delta_{\theta}(w) \right)} - \mathbbm{1}\{\sgn\{\delta_{\theta}(w) \} \neq 1\} \Big| \right] \]

Now let us again separate the integral form of the expectation into whether $\delta_\theta(w)$ is larger than some threshold value $\rho$:

\begin{align*}
  &\int_{w: |\delta_\theta(w)| < \rho } \Big|  \frac{1}{1 + \exp\left(k  \delta_{\theta}(w) \right)} - \mathbbm{1}\{\sgn\{\delta_{\theta}(w) \} \neq 1\} \Big| p(w) dw  \\
&+ \int_{w: |\delta_\theta(w)|> \rho}\Big|  \frac{1}{1 + \exp\left(k  \delta_{\theta}(w) \right)} - \mathbbm{1}\{\sgn\{\delta_{\theta}(w) \} \neq 1\} \Big| p(w) dw \\
\leq & \frac{\mathbb{P}\left(|\delta_\theta(w)| < \rho \right)}{2}\\
&+ \numberthis  \label{eq:int}\int_{w: |\delta_\theta(w)|> \rho}\Big|  \frac{1}{1 + \exp\left(k  \delta_{\theta}(w) \right)} - \mathbbm{1}\{\sgn\{\delta_{\theta}(w) \} \neq 1\} \Big| p(w) dw
\end{align*}

Then for \eqref{eq:int}, let us solve for $k$ such that the integrand is less than some fixed $\epsilon$. Due to symmetry about $x=0$ and $y=\frac{1}{2}$ of both the exponential and sign functions, without loss of generality we focus on when $x \geq 0$, i.e. we care about bounding $ \frac{1}{1 + \exp\left(k  \delta_{\theta}(w) \right)}$ from 0.

Then it suffices to find $k$ such that 

\begin{equation} \frac{1}{1 + \exp\left(k \rho \right)} \leq \epsilon, \label{det} \end{equation}
which upon rearranging yields 

\[ k \geq \frac{\log \frac{1}{\epsilon} - 1}{\rho}. \]

Therefore we choose $\epsilon \in o(1)$ (as a function of $n$), and appeal to assumption \ref{assump:3}.

Now let us go back to the first term of the RHS of \eqref{eq:tri1}. 

We have shown that $\|r - s_k\|_\infty \leq \epsilon + o(1) $. In light of this, we show that $\|  r (\theta^*) - s_{k}(\theta^s)\} \leq \epsilon + o(1).$

We have that $ s_{k}(\theta^s) \leq s_k(\theta^*) \leq r(\theta^*) + \epsilon + o(1) $
and symmetrically
$ r(\theta^*) \leq r(\theta^s) \leq s_k(\theta^s) + \epsilon + o(1).$ Combining the two yields 

\[\hat{s}_{n,k}(\hat{\theta}_n^s) - \epsilon - o(1) \leq \hat{r}_{n}(\hat{\theta}_n) \leq \hat{s}_{n,k}(\hat{\theta}^s_n) + \epsilon + o(1),\]
as desired.

Therefore we have bounded the term by $\epsilon + o(1)$.

All that remains is to bound the second term of \eqref{eq:tri1}. We can use the same argument as above, but in high probability. With probability at least $ 1- N_\Theta(\alpha) \exp\left(-2nt^2/\Delta^2 \right)$,  $\|s_k - \hat{s}_{n,k}\|_\infty \leq t $. By the same argument as above, then with probability exponentially approaching 1,

  \[ \|s_k(\theta^s) - \hat{s}_{n,k}(\hat{\theta}^s_n) \|  \leq t.\]

So our overall bound for the regret is $k\alpha + 2t + 2\epsilon + o(1)$ which holds with probability 
at least $ 1- N_\Theta(\alpha) \exp\left(-2nt^2/\Delta^2 \right).$

It remains to choose $\alpha, \epsilon, \rho, k, t$.

Let us choose $\alpha = n^{-1/3}$, so that $N_{\Theta}(\alpha) \in O(\exp(n^{2/3}))$ for SVMs and neural networks and $t = n^{-1/12}$, so that exponential probability $O(\exp(-n^{1/6}))$ is still achieved.

To choose $k$, we let $\rho = n^{-1/4}$ and $\epsilon = n^{-1}$, so that choosing $k = n^{1/4} \log n$ yields $k\alpha \in o(1)$ and $\epsilon \in o(1)$.

This yields probability exponentially approaching 1 that the bound is at most something $o(1)$.

\end{proof}

\subsection{Proof of Proposition 6.1}
\begin{proof}
We have that 

\begin{align} \hat{V}(\pi) &=  \frac{\sum_{i=1}^n \mathbbm{1}\{\pi(w_i) = x_i\} y_i }{\sum_{i=1}^n  \mathbbm{1}\{\pi(w_i) = x_i\}} \nonumber \\
&=  \frac{\frac{1}{n}\sum_{i=1}^n \mathbbm{1}\{\pi(w_i) = x_i\} y_i }{\frac{1}{n} \sum_{i=1}^n \mathbbm{1}\{\pi(w_i) = x_i\}} \label{eq:frac}.
\end{align}
Now consider the denominator to Equation \eqref{eq:frac}. Note that since each $x_i$ was chosen uniformly at random, we can express the event that $\pi(w_i) = x_i$ as the union of disjoint events $(\pi(w_i) = x) \cap (x_i = x)$, where the union is over the space of actions (articles) $x = 0, 1$ (recall we narrow down from 20 articles to 2 to yield a binary decision problem). Then since the event $x_i = x$ is independent of $\pi$, we can write

\begin{align*} \mathbb{P}(\pi(w_i) = x_i) &= \sum_{x=0}^{1} \mathbb{P} \left(\pi(w_i) = x \right) \mathbb{P} \left( x_i = x\right) \\
&= \frac{1}{2} \sum_{x=0}^{1} \mathbb{P} \left(\pi(w_i) = x \right) \\
&= \frac{1}{2}.
\end{align*}

Therefore we can apply the (weak) law of large numbers on each $\mathbbm{1}\{\pi(w_i) = x_i\}$, since they are independent and identically distributed Bernoulli random variables with $p = \frac{1}{2}$. So we have that \begin{equation} \frac{1}{n} \sum_{i=1}^n \mathbbm{1}\{\pi(w_i) = x_i\} \overset{p}{\to} \frac{1}{2}. \label{eq:wlln} \end{equation}

Now consider the inverse-propensity score (IPS) estimator, 
\[ \hat{V}_{IPS}(\pi) := \frac{\sum_{i=1}^n \mathbbm{1}\{\pi(w_i) = x_i\} y_i }{\mathbb{P}(\mu(w_i) = x_i)},\]
where $\mu$ is the uniform sampling policy.

From \cite{agarwal2019lec}, this estimator is unbiased for $V(\pi)$ and has variance which scales as $O(\frac{1}{n}).$ Therefore by Chebyshev's inequality,

\[ \mathbb{P} \left(| \hat{V}_{IPS}(\pi) - V(\pi) | \geq \epsilon \right) \leq \frac{\mathrm{Var}(\hat{V}_{IPS}(\pi))}{\epsilon^2} \overset{n \to \infty}{\to} 0, \]

and we conclude that $\hat{V}_{IPS}(\pi)$ is consistent for $V(\pi)$.

Since $\mathbb{P}(\mu(w_i) = x_i) = \frac{1}{2}$, it follows that the numerator for Equation \eqref{eq:frac} is consistent for $\frac{V(\pi)}{2}.$ We can finally combine this with Equation \ref{eq:wlln} and apply Slutzky's theorem to conclude that $\hat{V}(\pi)$ is consistent for $V(\pi)$. 
\end{proof}

\section{Experimental Setup}

The IHDP dataset was run on a M1 Pro Macbook Pro with 16GB RAM. The news recommender dataset was run on a cluster allocated 1 Intel(R) Xeon(R) Silver 4214R CPU @ 2.40GHz core, 1 RTX 3090 GPU, and 64GB RAM. 

To ensure reproducibility, random seeds were set for both \texttt{numpy} and \texttt{torch}.

\section{Code}

Here is an anonymized repository which contains our code: \url{https://anonymous.4open.science/r/ESR-4605/README.md}

\end{document}